\documentclass{article}

\usepackage[nonatbib, final]{neurips_2024}

\usepackage[utf8]{inputenc} % allow utf-8 input
\usepackage[T1]{fontenc}    % use 8-bit T1 fonts
\usepackage{url}            % simple URL typesetting
\usepackage{booktabs}       % professional-quality tables
\usepackage{amsfonts}       % blackboard math symbols
\usepackage{nicefrac}       % compact symbols for 1/2, etc.
\usepackage{microtype}      % microtypography
\usepackage{xcolor}         % colors

\usepackage{auxlib}

\title{Open-Book Neural Algorithmic Reasoning}

\author{
Hefei Li, \quad 
Chao Peng\thanks{Correspondence to Chao Peng and Chenyang Xu. The authors are ordered alphabetically. }, \quad  Chenyang Xu\footnotemark[1], \quad Zhengfeng Yang\\
Shanghai Key Laboratory of Trustworthy Computing \\ Software Engineering Institute\\ East China Normal University, Shanghai, China\\
\texttt
51255902127@stu.ecnu.edu.cn, \\
\{cpeng, cyxu, zfyang\}@sei.ecnu.edu.cn
}

\begin{document}

\maketitle

\begin{abstract}
Neural algorithmic reasoning is an emerging area of machine learning that focuses on building neural networks capable of solving complex algorithmic tasks. Recent advancements predominantly follow the standard supervised learning paradigm -- feeding an individual problem instance into the network each time and training it to approximate the execution steps of a classical algorithm. We challenge this mode and propose a novel open-book learning framework. In this framework, whether during training or testing, the network can access and utilize all instances in the training dataset when reasoning for a given instance.

Empirical evaluation is conducted on the challenging CLRS Algorithmic Reasoning Benchmark, which consists of 30 diverse algorithmic tasks. Our open-book learning framework exhibits a significant enhancement in neural reasoning capabilities. Further, we notice that there is recent literature suggesting that multi-task training on CLRS can improve the reasoning accuracy of certain tasks, implying intrinsic connections between different algorithmic tasks. We delve into this direction via the open-book framework. When the network reasons for a specific task, we enable it to aggregate information from training instances of other tasks in an attention-based manner. We show that this open-book attention mechanism offers insights into the inherent relationships among various tasks in the benchmark and provides a robust tool for interpretable multi-task training. 
\end{abstract}

\section{Introduction}\label{sec:intro}

Deep neural networks have achieved remarkable advancements in various areas, such as image processing~\cite{DBLP:journals/ijon/LiuWLZLA17,DBLP:conf/bigdata2/DengLGZY23} and natural language processing~\cite{DBLP:journals/ijon/LauriolaLA22,DBLP:journals/eswa/NamJ24}. In recent years, as deep learning continues to evolve, there has been an increasing desire to see deep neural networks take on more complex tasks. Algorithmic reasoning tasks~\cite{DBLP:journals/patterns/VelickovicB21,DBLP:conf/nips/DeacVMBTN21,DBLP:conf/log/VelickovicBKLHP22} have emerged as a particularly crucial category. In classical domains, deep neural networks have demonstrated their ability to learn predictive patterns from training data. The aspiration now is to extend this capability to the field of algorithmic reasoning, which motivates a burgeoning domain --- \emph{Neural Algorithmic Reasoning} (NAR).

Neural algorithmic reasoning was initially coined by~\cite{DBLP:conf/iclr/VelickovicYPHB20}. The central objective of this domain is to develop and train neural networks with the capability to imitate classical rule-based algorithms, such as sorting algorithms and graph algorithms. Networks built in this manner demonstrate the ability to perform algorithmic computations similar to traditional algorithms in reasoning tasks, while showcasing improved computational efficiency compared to them~\cite{DBLP:journals/corr/abs-2007-03629}. Moreover, recent literature~\cite{DBLP:conf/nips/XhonneuxDVT21,DBLP:conf/iclr/NumerosoBV23} shows that owing to the characteristics of deep learning, these networks exhibit flexibility in handling diverse input formats, making them robust even in scenarios where certain input features are missing.

%\xcy{There have been many recent advances in NAR~\cite{xxx}.}

\textbf{Challenging Benchmark for NAR.} CLRS Algorithmic Reasoning Benchmark proposed by~\cite{DBLP:conf/icml/VelickovicBBPBD22} is currently the most popular and definitive benchmark for evaluating the algorithmic capabilities of neural networks. This benchmark comprises 30 diverse algorithmic reasoning tasks extracted from the foundational algorithms textbook ``Introduction to Algorithms''~\cite{DBLP:books/daglib/0023376}, including sorting, searching, dynamic programming, graph algorithms, string algorithms, and more. Beyond the task diversity, another notable challenge of this benchmark is the \emph{significant differences} in scale between problem instances in the training and test sets. The test instances are substantially larger in scale compared to those in the training set.

% Notably,  the problem instances in the training set exhibit significant differences in scale (number of nodes, edges) compared to the test set, with the test instances being considerably larger than those in the training set.

There have been many recent advances in exploring CLRS~\cite{DBLP:conf/iclr/DiaoL23,DBLP:journals/tmlr/MahdaviSKHTL23,DBLP:conf/icml/BevilacquaNIBPB23,DBLP:conf/nips/DudzikV22,DBLP:conf/nips/RodionovP23,DBLP:journals/corr/abs-2406-09308}. As classical algorithms can often be represented by graph structures, several successful approaches leverage the Graph Neural Network (GNN) framework, including models such as PGN~\cite{DBLP:conf/nips/VelickovicBOPVB20} and MPNN~\cite{DBLP:conf/icml/GilmerSRVD17}. In addition to directly applying these classical GNNs, the literature has observed that the execution of some classical algorithms often relies on specific data structures. Consequently, there have been proposals to integrate classical GNNs with data structures like priority queues~\cite{npq} or stacks~\cite{recursive_ar} to enhance neural reasoning capabilities.

However, we notice that all prior approaches predict algorithmic executions based solely on their parameters and the features of a single input. Although this mode is commonly used in traditional supervised learning tasks~\cite{DBLP:conf/commnet/MrabetMF21,DBLP:journals/sncs/AljuaidA22},
it may not be well-suited for NAR due to the inherent difference between complicated reasoning tasks and traditional tasks like image processing. In practical scenarios, when recognizing images, extensive background knowledge is typically not required; but when faced with complex reasoning tasks, a substantial amount of background knowledge is often necessary to complete various aspects of the reasoning process. In such situations, having real-time illustrative examples or formulas available for reference can significantly reduce our memory burden, thereby enhancing task completion. This naturally raises a question:

%\vspace{5pt}
\begin{center}
    \begin{minipage}{0.8\textwidth}
    \begin{center}
    \emph{If allowing a neural network to access additional examples for reference during reasoning, will its reasoning capability improve as a result?}
\end{center}
\end{minipage}
\end{center}

\subsection{Our Contributions}

We explore the aforementioned question and introduce open-book neural algorithmic reasoning. In this model, the neural architecture is enhanced with an additional memory component that stores representations of instances in the training dataset. Whether during training or testing, whenever the network engages in reasoning for a specific instance, it has the capability to leverage this supplementary memory to aggregate information from other instances within the training set, akin to an open-book exam. The main results of the paper are summarized as follows:

\begin{itemize}
    \item We present a general framework for open-book NAR. This framework builds upon the foundation of previous NAR architectures by introducing two additional modules for embedding and information aggregation from the training set, and can seamlessly integrate with existing methods. 
    We further provide a detailed implementation of the framework, which is grounded in the cross-attention mechanism. This design not only caters to single-task training but also proves to be highly effective in scenarios involving multi-task training.%We provide a detailed implementation of the open-book framework. The implementation is based on the cross-attention mechanism  
    
    \item Empirical evaluations are conducted on the challenging CLRS Benchmark~\cite{DBLP:conf/icml/VelickovicBBPBD22}.
    We incorporate the proposed framework with three popular network architectures in the literature. The results demonstrate that each architecture's reasoning capability can be improved significantly when utilizing the training instances through the framework. Across the majority of the reasoning tasks within the benchmark, the framework yields state-of-the-art results. 
    
    \item Multi-task training is also investigated in the paper. As highlighted in~\cite{DBLP:conf/log/IbarzKPNBCDBVRD22}, on certain reasoning tasks, a generalist network trained on all datasets in CLRS outperforms the networks trained in a single-task manner. We provide an interpretation of this observation using the proposed open-book framework. Specifically, when training a neural network to solve a task, we input information from other task datasets into the framework for its use. The results show that our open-book framework can nearly replicate the effects of multi-task training for each algorithmic task, while in some tasks, it even achieves higher accuracies. Additionally, our attention-based implementation enables us to analyze the attention weights of various tasks, facilitating a deeper understanding of the intrinsic relationships among tasks. A ``paired training'' experiment is further conducted to verify the effectiveness of the learned attention weights. 
\end{itemize}

\subsection{Other Related Work}

Our work is closely aligned with the exploration of non-parametric models~\cite{DBLP:conf/ac/Rasmussen03,DBLP:conf/cvpr/ShuaiWZWZ15,DBLP:conf/nips/JiangDXLL22}, where models abstain from training specific parameters and, instead, utilize dependencies among training data points for predictions. Our framework can be viewed as a fusion of deep neural networks and non-parametric models. We have noted analogous efforts in recent work within the field of image processing~\cite{DBLP:conf/nips/KossenBLGRG21}. This work focuses on the CIFAR-10 dataset, employing self-attention mechanisms among different points in the dataset to finish image classification tasks.

%\subsection{Paper Organization}

% The paper is structured as follows. In~\cref{sec:pre}, we introduce the background of neural algorithmic reasoning. Following that, in~\cref{sec:method}, we present our open-book framework and offer a detailed implementation of the framework.~\cref{sec:exp} is dedicated to the experimental section, where we conduct empirical evaluations on the CLRS benchmark. Lastly,~\cref{sec:con} concludes the paper.

\section{Preliminaries}
\label{sec:pre} 

This section introduces the setting of an NAR dataset formally and outlines the standard paradigm employed in NAR.

\textbf{NAR Dataset.} The objective of an NAR task is to train a neural network such that it can imitate each execution step of a classical algorithm on given problem instances. Hence, a NAR dataset is labeled by a specific problem and the algorithm employed to solve it. Each data point includes a problem instance, represented by a graph structure, and the corresponding algorithm execution on that instance, conveyed through a sequence of graph-structured states. Denote by $\vx$ the problem instance and by $\vy = \{\vy^{(1)},...,\vy^{(t)},... \}$ the algorithm execution, where $\vy^{(t)}$ signifies the graph-structured states (e.g., the current nodes in the queue of breadth-first search) at the $t$-th step of the algorithm. 

\textbf{Training Objective.} The training objective of the neural network is to perform sequential reasoning tasks over a given problem instance. At each step $t$, the network takes as input the pair $\left( \vx, \vy^{(t-1)} \right)$ and produces the output $\vy^{(t)}$. This process enables the neural network to learn and predict the evolution of the algorithmic execution on the problem instance in a step-wise fashion.

\textbf{Encode-Processor-Decode Paradigm.} To achieve the aforementioned step-wise objective, the literature follows the standard \emph{encode-process-decode} paradigm~\cite{DBLP:conf/cogsci/HamrickABZMTB18}, which consists of three modules: Encoder, Processor, and Decoder. At each step $t$, the input $ \left( \vx, \vy^{(t-1)} \right) $ traverses through these modules sequentially\footnote{For simplicity, we abuse the notion slightly, allowing $\vy^{(t-1)}$ to represent the outcome of the last step.}:
\begin{itemize}
    \item The encoder module encompasses multiple neural networks that operate on $ \left( \vx, \vy^{(t-1)} \right) $, thereby transforming it into a collection of graph-structured hidden states. Use $G=(V, E)$ to denote the graph structure. Following this module, we obtain $h_{v}$ corresponding to each node $v\in V$, $h_{vu}$ associated with each edge $(v,u)\in E$, and $h_g$ representing the hidden state of the entire graph $G$.
    %The encoder module contains several neural networks, transforming $ \left( \vX, \vY^{(t-1)} \right) $ into a set of graph-structured hidden states. After this module, we have $h_{i}$ for each node $i$, $h_{ij}$ for each node pair $(i,j)$ and $h_g$ denoting the hidden state of the whole graph. 

    \item The processor module usually consists of a graph neural network. This module maintains the historical hidden states of nodes, edges, and the graph: $\{h_v^{(t-1)}\}_{v\in V}$, $ \{h_{vu}^{(t-1)}\}_{(v,u)\in E}$, $ h_g^{(t-1)}$, and integrate them with the newly generated states $\{h_v\}$, $ \{h_{v,u}\}$, $ h_g$ to yield updated states. We borrow the language of the message-passing architecture~\cite{DBLP:conf/icml/GilmerSRVD17} to formalize this process.  For brevity, the following focuses only on updating the state of each node $v$. At each step $t$, the node computes and aggregates messages $m_{uv}$ from its incoming edges, updating its own hidden state:
    \begin{equation*}
        \begin{aligned}
            z_v^{(t)} & \leftarrow \vf_1 \left( h_v, h_{v}^{(t-1)} \right)\; ; & m_{uv} & \leftarrow \vf_2\left(z_v^{(t)}, z_u^{(t)}, h_{uv},h_g \right) \;\;\; \forall (u,v) \in E \; ; \\
            M_v &\leftarrow \bigoplus\limits_{u:(u,v)\in E} m_{uv} \; ; & h_v^{(t)} & \leftarrow \vf_3 \left(z_v^{(t)}, M_v\right).
        \end{aligned}
    \end{equation*}
    Different processors employ different layers $\vf_1$, $\vf_2$, $\vf_3$, and aggregation function $\bigoplus$. 

    \item The decoder module utilizes the states $\vh^{(t)}$ as input to forecast the algorithmic execution $\vy^{(t)}$ at step $t$. It is noteworthy that recent literature~\cite{DBLP:conf/log/IbarzKPNBCDBVRD22} also incorporates $\vx$ and $\vy^{(t-1)}$ within this module.
    %The decoder module takes as input the states $\vH^{(t)}$ to predict the algorithm execution $\vY^{(t)}$ at step $t$. It is worth noting that recent literature~\cite{xxx} additionally makes use of $vX$ and $\vY^{(t-1)}$ in this module.  
\end{itemize}

\section{Open-Book Reasoning}\label{sec:method}

The paradigm above can be denoted by a function $\cF$ mapping $\vx$ to $\vy$ for each data point. Given a NAR dataset, this function implies a standard supervised learning mode: during a training step, a (or a mini-batch of) random datapoint $(\vx, \vy)$ is selected. The loss between $\cF(\vx)$ and $\vy$ is then computed, and the parameters in $\cF$ are updated accordingly. In this section, we go beyond the individual $ \vx \rightarrow \vy $ mode in conventional supervised learning, exploring a more general and practical learning paradigm.

\subsection{Framework}

We introduce an open-book reasoning framework. Within the framework, when the network is tasked with solving problem instance $\vx$ and deducing $\vy$, it not only utilizes $\vx$ as input but is also allowed to leverage information from other data points within the training set during the reasoning process. 
%The intuition behind is that when we do reasoning 

% \begin{wrapfigure}{r}{0.65\linewidth}
%     \centering
%   \includegraphics[width=0.9\linewidth]{Figures/framework}
%   \caption{An illustration of the open-book framework. At each reasoning step $t$, we simultaneously input $(\vx,\vy^{(t-1)})$ and instances from the training set $\vT$, yielding $\vy^{(t)}$.}
%   \label{fig:framework}
% \end{wrapfigure}

\begin{figure}[ht]
    \centering
  \includegraphics[width=0.8\linewidth]{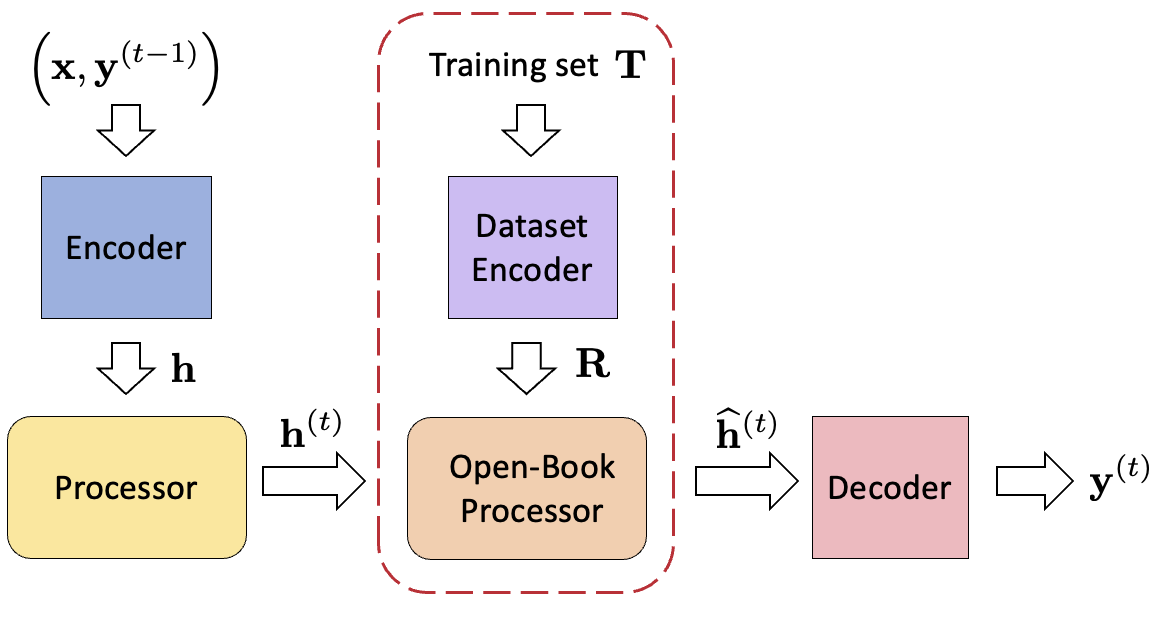}
  \caption{An illustration of the open-book framework. At each reasoning step $t$, we simultaneously input $(\vx,\vy^{(t-1)})$ and instances from the training set $\vT$, yielding $\vy^{(t)}$.}
  \label{fig:framework}
\end{figure}

The intuition behind the open-book framework is analogous to our real-world scenario of solving algorithmic problems or engaging in other reasoning tasks. In practical situations, we often consult textbooks and refer to example problems to aid in completing tasks. Typically, the structure and solutions of these examples provide substantial assistance in our reasoning process.
Denoting the training set as $\vT$, the framework essentially aims to learn a comprehensive function $\cF: \vx \cup \vT \rightarrow \vy $.

% \begin{figure*}[h]
%   \centering
%   \includegraphics[width=0.9\linewidth]{Figures/framework}
%   \caption{An illustration of the open-book framework. At each reasoning step $t$, we simultaneously input $(\vx,\vy^{(t-1)})$ and instances from the training set $\vT$, yielding $\vy^{(t)}$.}
%   \label{fig:framework}
% \end{figure*}

An illustration of the framework is present in~\cref{fig:framework}. In addition to the original three modules, we introduce two new modules: \emph{Dataset Encoder} and \emph{Open-Book Processor}:
\begin{itemize}
    \item The dataset encoder module employs an encoding function~$f_E$ to compute the latent feature of each data point $\vd_i = (\vx_i,\vy_i)$ in the training set: $\vr_i \leftarrow \vf_E\left( \vx_i,\vy_i  \right).$
    % \begin{equation*}
    %     \begin{aligned}
    %         \vr_i \leftarrow \vf_E\left( \vx_i,\vy_i  \right).
    %     \end{aligned}
    % \end{equation*}
    It is worth noting that this encoder module is essentially different from the original one. It maps an entire data point $\vd_i=(\vx_i,\vy_i)$, encompassing the ground truth of each node (and edge) at each step, into a single representation $\vr_i$.

    \item The open-book processor module is incorporated between the original processor and decoder modules. The output $\vh^{(t)}$ from the processor no longer directly feeds into the decoder; instead, it passes through the open-book processor, where it undergoes information aggregation with the training data representation $\vR=\{\vr_1,...,\vr_i,... \}$ (generated by the dataset encoder). Subsequently, the open-book processor produces the latent features $\vhath^{(t)} $ required by the decoder. Formally, for each node $v\in V$, $\hath_v^{(t)}  \leftarrow \vf_P\left( h^{(t)}_v,\vR  \right).$
    % \begin{equation*}
    %     \begin{aligned}
    %         \hath_v^{(t)}  \leftarrow \vf_P\left( h^{(t)}_v,\vR  \right).
    %     \end{aligned}
    % \end{equation*}
    
    %The open-book processor module is introduced between the original processor and decoder modules. The output $\vh^t$ from the processor no longer directly feeds into the decoder; instead, it undergoes information aggregation through the open-book processor and the dataset representation generated by the dataset encoder. Subsequently, it generates the latent feature required by the decoder.
    
\end{itemize}

The central component of the framework is the open-book processor module. Within this module, the current problem instance to be solved is integrated with examples from the training set. It is crucial to acknowledge that this integration has both advantages and disadvantages. While it often enhances the architecture's reasoning capabilities, there are instances when it may lead to counterproductive effects, particularly during multi-task training. We will elaborate on this in the experiments.

%It is important to note that such an integration has its pros and cons, providing assistance in accurate reasoning but potentially introducing some level of interference. We will show this in the experimental section.%, we will analyze these aspects.

%The most crucial component of the framework is the open-book processor module.  

\subsection{Attention-Based Implementation}

Diverse implementations within the framework can be achieved by employing different functions for $\vf_E$ and $\vf_P$. For the ease of investigating multitask training, we adopt an attention mechanism-based implementation. A description of the network implementation and training is given in~\cref{alg:openbook}.
%To provide a clearer depiction of the entire operation of open-book learning, we give a comprehensive description of the network implementation and training in~\cref{alg:openbook}.

%To derive the architecture intuition, we 

\begin{algorithm*}[h]
\caption{Attention-Based Implementation of Open-Book Reasoning}
\label{alg:openbook}
\begin{algorithmic}[1]
\Require Training set $\vT$.
% \State  {\color{gray} /* ------------------------Training Phase------------------------ */}
\While{current epochs $\leq $ maximum training epochs}
\State Initialize the hidden state $\vh^{(0)}$ and $\vhath^{(0)}$.
\State Randomly pick a \emph{target} data point $\vd=(\vx,\vy) \in \vT$ and several \emph{auxiliary} data points $\vd_1,\vd_2,...\vd_{\ell}\in \vT$.
%\State \xcy{xxxcheckxxxlayernorm}
%\State %Set a linear layer $\linear (\cdot )$.
\Comment{\textbf{Dataset Encoder} }
\For{ each auxiliary data point $\vd_i=(\vx_i,\vy_i)$ }
\State Let $G=(V,E)$ represent the underlying graph of $\vd_i$. This data point encompasses a state sequence $<\vx_i=\vy_i^{(0)}, \vy_i^{(1)},...,\vy_i^{(t)},... >$ on the graph.%, where $\vy_i^{(t)} = \left(\{ y_{i,v}^{(t)}\}_{v\in V},\{ y_{i,e}^{(t)}\}_{e\in E},  y_{i,g}^{(t)} \right)$ indicates the node states, the edge states and the graph state at the $t$-th algorithm step.
\State Randomly select two adjacent states $\vy_i^{(p)}, \vy_i^{(p+1)} $ from the sequence.
\For{ each node $v\in V$ }
\State \parbox[t]{\dimexpr\linewidth-\algorithmicindent}{Let $y_v$ and $y_v'$ be node $v$'s states in $\vy_i^{(p)}$ and $\vy_i^{(p+1)}$ respectively.}
\State Employ a linear layer $ z_v \leftarrow  \linear\left(\frac{1}{2}\cdot (y_v+y_v')\right) $.
\EndFor
\State $\vr_i \leftarrow \frac{ 1}{|V|}\cdot \sum_{v\in V}z_v$.
\EndFor
\State Define $\vR := [ \vr_1,...,\vr_{\ell} ] $.
\For{ each algorithm execution step $t$ of the target data point}
\State Feed $\vx$ and the outcome from the previous step into the encoder and processor sequentially to obtain the hidden states $\vh^{(t)}$.
\State Employ a linear layer: $\vR^{(t)}\leftarrow \vR \;\Vert\; \linear (\vh^{(t)}) $. \Comment{\textbf{Open-Book Processor}} 
\State Set a QKV attention function with linear layers $\query(\cdot), \key(\cdot),\ovalue(\cdot) $.   
\State Compute a cross-attention between $\vh^{(t)}$ and $\vR^{(t)}$: \[\qquad \qquad \qquad \qquad \vhath^{(t)} \leftarrow \softmax\left(\frac{\query \left(\vh^{(t)}\right) \cdot \key\left( \vR^{(t)}\right) }{\sqrt{d_k}}\right) \;\cdot\; \ovalue\left(\vR^{(t)}\right),\] \qquad \qquad \quad where $d_k$ is the dimension of the key vectors.
\State Add a gate function: $\vhath^{(t)}\leftarrow \gate\left( \vh^{(t)}, \vhath^{(t)} \right).   $
\State Feed $\vhath^{(t)}$ into the decoder module, yielding the predictions of this step.
\EndFor
\State Compute the prediction loss and update the network parameters.
\EndWhile
% \State {\color{gray} /* ------------------------Testing Phase------------------------ */} 
\end{algorithmic}
\end{algorithm*}

From the description, a \emph{target} data point and several \emph{auxiliary} data points are randomly selected in each training iteration. The target data point serves as the focal point for neural optimization in this iteration: the network predicts its ground truths, computes the loss, and consequently updates the network parameters. The auxiliary data points assist the network in reasoning for the target data point. Their latent features are obtained through the dataset encoder, and they subsequently influence predictions through the open-book processor.

At each algorithmic step $t$, the hidden states $\vh^{(t)}$ are computed conventionally. However, these states are not directly input into the decoder. Instead, they need to undergo cross-attention with the representations of auxiliary data points within the open-book processor module. This design allows the network to incorporate hints provided by the auxiliary data points during the reasoning process.

The construction of the dataset encoder is a bit subtle. We observe a crucial aspect that all these pieces of information ultimately serve the decoder module. In a single algorithmic step, the decoder's role is to facilitate the transition between two adjacent states throughout the entire reasoning process. Therefore, to better provide effective hints to the final decoder, for each auxiliary data point, we randomly sample a pair of adjacent states from its corresponding state sequence. Subsequently, we employ a linear layer to yield the latent representations of these data points.

\textbf{Remark.} The testing process is essentially similar to the training. It is worth noting that during the testing phase, the target data points are sourced from the testing set, while the auxiliary data points must still originate from the training set.

\section{Experiments}\label{sec:exp}

%\xcy{Running time}
This section evaluates the open-book implementation empirically on the CLRS benchmark. We aim to investigate the following three questions during the experiments:

\begin{itemize}
    \item For various processor architectures present in the literature, can the open-book framework consistently enhance their empirical performances across the majority of the algorithmic tasks within the CLRS benchmark? 
    
    %a specific reasoning task, can the performance of a network on it be improved by feeding its training set into the open-book framework?

    \item There is a recent literature~\cite{DBLP:conf/log/IbarzKPNBCDBVRD22} proposing a multi-task training approach for CLRS. They train a common network for various tasks in the benchmark and find that some tasks benefit from the multi-task approach, achieving higher accuracy than when trained individually. In the context of the open-book setting, does this phenomenon imply that incorporating training sets from various tasks into the open-book framework may enhance the network's performance on certain tasks?
    
    % Given a set of reasoning tasks, 
    
    % we can either train individual networks for each task or employ a common network for multi-task training. If a reasoning task benefits from the multi-task approach (achieving higher accuracy than when trained individually), does this imply that incorporating training sets from other tasks into the open-book framework will enhance the network's performance on that task?
    
    %Suppose there exists a reasoning task benefiting from the multi-task manner. Does this suggest that the network's performance on that task can be enhanced by incorporating training sets from other tasks into the open-book framework?

    \item Can the attention-based implementation serve as a robust tool for interpretable multi-task training? When integrating training sets from various tasks into the open-book framework for a specific task, the network eventually learns attention weights in the open-book processor, signifying the task's relevance to other tasks. Does this imply that if a task performs better in multi-task training than in single-task training, retaining only those tasks with prominent attention for multi-task training can still outperform single-task training?  
\end{itemize}

% Our question?

To tackle these questions, we conduct three types of experiments\footnote{The codes are provided in \url{https://github.com/Hoferlee1/Open-Book}}: single-task augmenting, multi-task augmenting, and multi-task interpretation. Note that our ``multi-task augmenting'' experiment differs essentially from traditional multi-task training; here, we still train the network for a specific task, but with the inclusion of datasets from other tasks in the dataset encoder. Additional ablation experiments are also conducted (see the appendix). We initially outline the experimental setup and subsequently delve into each experiment. %individually.
%To address these questions, we conduct three sets of experiments: single-task training, multi-task training, and multi-task interpreting. We first describe the experimental setup and then introduce each experiment one by one.

\subsection{Setup}

\textbf{Baselines.} We incorporate the open-book framework into three existing processor architectures: PGN~\cite{DBLP:conf/nips/VelickovicBOPVB20}, MPNN~\cite{DBLP:conf/icml/GilmerSRVD17} and Triplet-GMPNN~\cite{DBLP:conf/log/IbarzKPNBCDBVRD22}.
% architecture~\cite{DBLP:conf/log/IbarzKPNBCDBVRD22}, leveraging its encoder, processor, and decoder components. 
Given that the feature dimension of hidden states is set to 128 in the literature, we adjust the parameters of the dataset encoder and open-book processor to ensure seamless integration. The results (F1 scores) achieved by open-book reasoning are compared with them. Moreover, we also compare the performance with other recent architectures like Memnet~\cite{DBLP:conf/icml/VelickovicBBPBD22} and NPQ~\cite{npq}.
% classical GNN architectures, such as Memnet~\cite{DBLP:conf/icml/VelickovicBBPBD22}, MPNN~\cite{DBLP:conf/icml/VelickovicBBPBD22}, and PGN~\cite{DBLP:conf/icml/VelickovicBBPBD22}.

% We integrate the open-book framework with the existing architecture Triplet-GMPNN~\cite{triplet}, borrowing its encoder, processor, and decoder. As the feature dimension of hidden states in Triplet-GMPNN is set to be 128, we also set 

% The two most crucial baselines we compare against are Triplet-GMPNN~\cite{triplet} and RT~\cite{rt}, which play a pivotal role in addressing the aforementioned questions. 

% Moreover, we include Memnet~\cite{memnet}, MPNN~\cite{mpnn}, PGN~\cite{pgn}, and NPQ~\cite{npq} in experiments as important references for comparison.

\textbf{Computational Details.} %In our implementations, $\vf_{\linear}$, $\vf_{\linear\textendash 1}$ and $\vf_{\linear\textendash 2}$ are individual linear layers.
The experiments are conducted on a machine equipped with an i7-13700K CPU, an RTX 4090 GPU, and an RTX A6000 GPU. The results are averaged over 4 runs.
To ensure fair comparisons, we follow the widely-used experimental hyperparameter settings in~\cite{DBLP:conf/log/IbarzKPNBCDBVRD22}, where the batch size is 32 and the network is trained for 10,000 steps by Adam optimizer with a learning rate of 0.001. During each training and testing iteration, we allow~\cref{alg:openbook} to sample $240$ auxiliary data points and use only one attention head. The average training time for each reasoning task is approximately 0.5 GPU hours.

 % while for CEF-RT, we set the batch size to 4 and the network is trained for 10,000 steps by Adam optimizer with a learning rate of 0.00025.  We remark that with the introduction of a preprocessor module in our framework, the average training time increases by 3 to 5 minutes for both methods. This increment represents less than 10\% of the total runtime, indicating the efficiency of our implementations. More details are deferred to the appendix.

%To answer the aforementioned three questions, we conduct three types of experiments. 

\subsection{Single-Task Augmenting}\label{sec:single_training}

\begin{figure*}[tb]
  \centering
  \includegraphics[width=\linewidth]{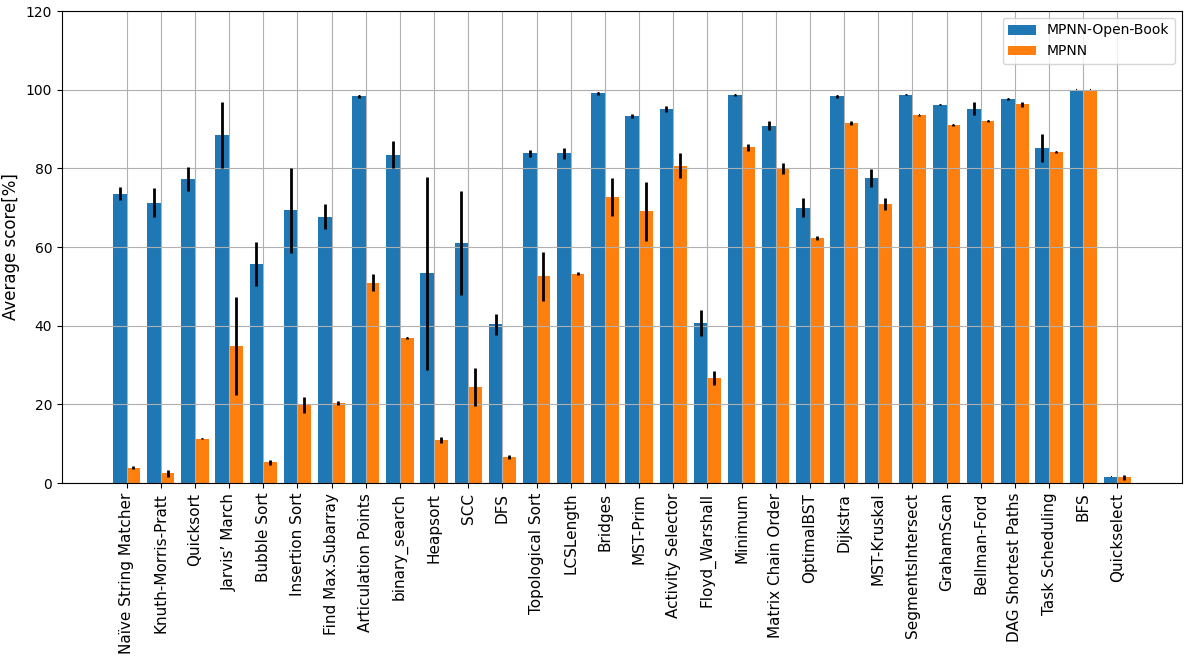}
  \caption{Comparison of the MPNN architecture's performance before and after augmentation with the open-book framework. The 30 tasks are arranged in descending order of improvement magnitude.}
  \label{fig:MPNN}
\end{figure*}

\begin{table*}[h]
    \centering
    \caption{The summary of our results on each task category in CLRS. The best-performing results in each row are highlighted in bold. To save space, we use the column ``Prior Best'' to denote the best results among four existing approaches: Memnet~\protect\cite{DBLP:conf/icml/VelickovicBBPBD22}, PGN~\protect\cite{DBLP:conf/icml/VelickovicBBPBD22}, MPNN~\protect\cite{DBLP:conf/icml/VelickovicBBPBD22}, and NPQ~\protect\cite{npq}, and the column ``Ours'' to denote the best results achieved by applying the open-book framework to the three existing architectures.}
    \label{tab:8sort}
    \begin{tabular}{cccc}
    \hline
        Task Category &
  %\begin{tabular}[c]{@{}c@{}}Prior Best\end{tabular} &
  \begin{tabular}[c]{@{}c@{}}Prior Best\end{tabular} &
  \begin{tabular}[c]{@{}c@{}}Triplet-GMPNN\end{tabular} &
  \multicolumn{1}{c}{\begin{tabular}[c]{@{}c@{}}Ours\end{tabular}}  \\ \hline
        Graphs   & 64.98\%±2.59 & 81.41\%±1.53 & \textbf{85.37\%±1.73}   \\ 
        Geometry   & 92.48\%±1.35 &94.09\%±0.77  & \textbf{96.55\%±0.50}   \\ 
        Strings   & 4.08\%±0.57 & 49.09\%±4.78  & \textbf{72.41\%±2.66}     \\ 
        Dynamic Programming   & 76\%±2.47 & 81.99\%±1.30  &\textbf{82.14\%±1.45}    \\ 
        Divide and Conquer   & 65.23\%±2.56  & \textbf{76.36\%±0.43}  & 74.52\%±1.88    \\ 
        Greedy  & 84.13\%±2.59  & 91.22\%±0.40  & \textbf{93.40\%±2.12}   \\ 
        Search   & 56.11\%±0.36 & 58.61\%±1.05  & \textbf{63.15\%±0.90}     \\
        Sorting   & 71.53\%±0.97 & 60.38\%±5.25  & \textbf{83.65\%±3.06}    \\ 
          \hline
        % \textgreater 90\% & 7/30 & 11/30 & a & b & c  \\ 
        % \textgreater 80\% & 11/30 & 17/30 & a & b & c  \\ 
        % \textgreater 60\% & 23/30 & 24/30 & a & b & c  \\ \hline
    \end{tabular}
\end{table*}

This subsection considers a single-task environment: for each reasoning task in CLRS, both target and auxiliary data points in~\cref{alg:openbook} are sourced from its own dataset. 
We create comparison charts for results on three existing architectures. We present one chart~\cref{fig:MPNN} in the main body, while the other two are deferred to~\cref{sec:add_single}.%~\cref{sec:add_single}.
%The results are shown in~\cref{fig:single-task}. 
The figure uses bar charts to illustrate average scores for each task, with standard deviations denoted by black lines. Additionally, we arrange the tasks in descending order of improvement magnitude to better illustrate trends.

We also provide tables to comprehensively compare the accuracies that the open-book framework yields with existing results. In CLRS, the 30 tasks are partitioned into 8 categories: Divide and Conquer, Dynamic Programming, Geometry, Graphs, Greedy, Search, Sorting, and Strings. So we present two tables (\cref{tab:results} and~\cref{tab:8sort}): one showcasing the performance on the 30 individual tasks and another displaying the average performance for each of the 8 task categories. 
% Due to space constraints, the latter is included in the main body (\cref{tab:8sort}), while the former is deferred to~\cref{tab:results}. %the full version of this paper. %~\cref{sec:add_single}.

%We present two tables to showcase the performance on the 30 individual tasks and the average performance for each of the 8 task categories.

From the figures and tables, we observe that our approach outperforms the original architectures in the majority of tasks. The improvements provided by the open-book framework are particularly significant for certain tasks, such as the Naive String Matcher task (see~\cref{fig:MPNN}). However, we also notice a relatively large standard deviation in performance for some tasks. We attribute this variability to the fact that during testing, we sample data from the training set and input it into the dataset encoder each time. The quality of the sampled data influences the final inference results, leading to performance fluctuations.

%\xcy{one figure. Triplet-GMPNN and our ,}

\subsection{Multi-Task Augmenting}\label{sec:multi_training}

This subsection considers a ``multi-task'' environment: for each task in CLRS,~\cref{alg:openbook} selects target points from its own dataset, while the sampled auxiliary points are drawn from all datasets in CLRS. Since CLRS comprises 30 datasets, in each iteration, we randomly sample 8 instances from each dataset, ensuring that the total number of auxiliary points remains the same as in the single-task experiment, i.e., 240.
Given that Triplet-GMPNN is the only architecture used for multi-task training in the literature, both this subsection and the following one ``multi-task interpretation'' focus exclusively on the results obtained by integrating the open-book framework with Triplet-GMPNN.

\begin{figure*}[tb]
  \centering
  \includegraphics[width=\linewidth]{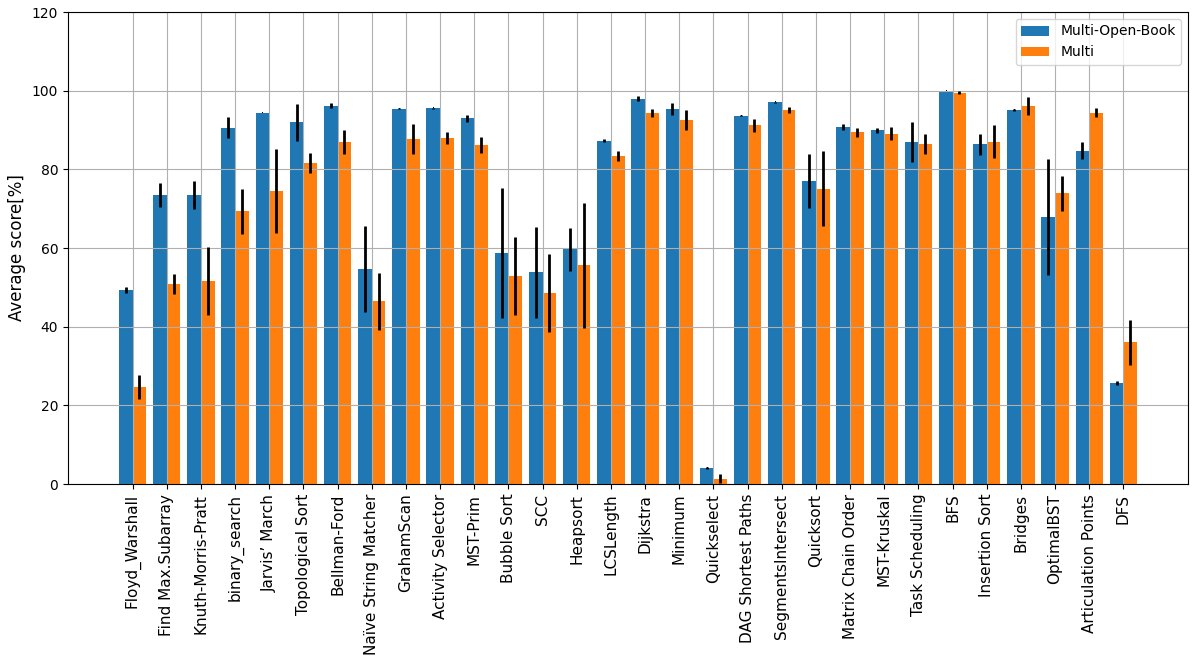}
  \caption{Comparisons between our multi-task augmented approach and Triplet-GMPNN. The 30 tasks are arranged in descending order of improvement magnitude.}
  \label{fig:multi-task}
\end{figure*}

The results are present in~\cref{fig:multi-task}. We find that incorporating data from different tasks into the open-book processor indeed replicates multi-task training. Our multi-task augmented method closely matches the previous multi-task training results, and even outperforms them on the vast majority of tasks.
It is worth noting that multi-task training requires simultaneous training on all 30 algorithmic tasks, which is extremely time-consuming. If the goal is simply to enhance performance on a specific task using multi-task training, the cost is substantial. However, with the open-book framework, we can nearly achieve the effects of multi-task training on a target task in approximately the same amount of time it takes to train a single algorithm.

% We show that the multi-task

% It is evident that, compared to the traditional Triplet-GMPNN, our approach outperforms in the majority of tasks. For the tasks that exhibit improvement with multi-task training in the literature, our multi-task augmenting method also yields better results than single-task training. Moreover, for tasks that show a decline in performance during multi-task training, our method demonstrates robustness, with performance comparable to or even surpassing single-task training results in some tasks.

%\xcy{One figure with single-task multi-task 30-bank descending over the single-task }
%This subsection considers a ``multi-task'' environment: for each reasoning task in CLRS,~\cref{alg:openbook} samples target points from its own dataset while the sampled auxiliary points are from all datasets in CLRS. As CLRS involves 30 datasets, in each iteration, we randomly sample 8 instances for each task. 

\begin{table*}[t]
\caption{For each target (task), we show the task with the highest attention weight among other tasks in column ``Auxiliary''. We use bold text to indicate when the paired tasks belong to the same algorithmic category.}
\label{tab:attention}
\begin{subtable}{0.5\linewidth}
\centering
\begin{tabular}{cc}
\hline
  \begin{tabular}[c]{@{}c@{}}Target\end{tabular} &
  \begin{tabular}[c]{@{}c@{}}Auxiliary \end{tabular}  \\ \hline
 Activity Selector & Topological Sort  \\ %\hline
 Articulation Points & Knuth-Morris-Pratt \\ %\hline
Bellman-Ford & Bridges \\ %\hline
 BFS & Task Scheduling \\ %\hline
 \textbf{Binary Search} & \textbf{Quickselect} \\ %\hline
 Bridges & Optimal BST  \\ %\hline
Bubble Sort & Task Scheduling \\ %\hline
 DAG Shortest Paths & Naïve String Matcher  \\ %\hline
DFS & Binary Search  \\ %\hline
\textbf{Dijkstra} & \textbf{Bellman-Ford}  \\ %\hline
Find Max. Subarray & Jarvis’ March \\ %\hline
Floyd-Warshall & Heapsort \\ %\hline
Graham Scan & Quicksort  \\ %\hline
Heapsort & Activity Selector  \\ %\hline
Insertion Sort & Minimum  \\ \hline
\end{tabular}
\end{subtable}
\begin{subtable}[t]{0.5\linewidth}
\centering
\begin{tabular}{cc}
\hline
  \begin{tabular}[c]{@{}c@{}}Target \end{tabular} &
  \begin{tabular}[c]{@{}c@{}}Auxiliary \end{tabular} \\ \hline
Jarvis’ March & MST-Kruskal  \\ %\hline
Knuth-Morris-Pratt & Quicksort  \\ %\hline
LCS Length & Dijkstra  \\ %\hline
Matrix Chain Order & Jarvis’ March \\ %\hline
Minimum & Quicksort  \\ %\hline
MST-Kruskal & Heapsort  \\ %\hline
 \textbf{MST-Prim} & \textbf{Bridges}   \\ %\hline
Naïve String Matcher & LCS Length  \\ %\hline
Optimal BST & Find Max. Subarray  \\ %\hline
Quickselect & Dijkstra \\ %\hline
Quicksort & BFS \\ %\hline
Segments Intersect & Topological Sort  \\ %\hline
SCC& Task Scheduling \\ %\hline
Task Scheduling& Heapsort \\ %\hline
\textbf{Topological Sort} & \textbf{DAG Shortest Paths} \\ \hline
\end{tabular}
\end{subtable}
\end{table*}

\subsection{Multi-Task Interpretation}\label{sec:multi_interp}

%\xcy{Here}
This subsection delves into interpreting multi-task training. In our multi-task augmenting experiments, the acquired attention weights in the open-book processor reveal the significance of each task in relation to others. Specifically, for each task, we aggregate the attention weights of each node at every algorithmic step on each test instance. The resulting 30-dimensional vector is then normalized, serving as the total attention vector for that task relative to other tasks in the benchmark.~\cref{tab:attention} shows the task with the highest attention weight for each task. Moreover, we present a heatmap regarding the attention weights among CLRS tasks in~\cref{sec:add_multi}.

\begin{table*}[tb]
\caption{Comparisons among three training manners under Triplet-GMPNN.}
\label{tab:attention_verify}
\centering
\begin{tabular}{cccc}
\hline
Task               & Single-Task & Multi-Task & Paired-Task \\ \hline
Heapsort           &             31.04\%±5.82&          \textbf{55.62\%}±15.91&         46.63\%±10.43
\\ %\hline
Knuth-Morris-Pratt &             19.51\%±4.57&           51.61\%±8.63&         \textbf{65.67\%±12.36}
\\ %\hline
 Insertion Sort
& 78.14\%±4.64
& 87.00\%±4.16
&\textbf{95.78\%±0.80}
\\  %\hline
 LCS Length
& 80.51\%±1.84
& 83.43\%±1.19
&\textbf{85.86\%±1.47}
\\  %\hline
 Quicksort
& 64.64\%±5.12
& 75.10\%±9.52
&\textbf{88.43\%±6.25}
\\  %\hline
 SCC
& 43.43\%±3.15
& 48.48\%±9.96
&\textbf{73.39\%±3.00}
\\  %\hline
 Jarvis’March
& 91.01\%±1.30
& 74.51\%±10.71
&\textbf{94.44\%±0.63}
\\  %\hline
 MST-Kruskal
& 89.80\%±0.77
& 89.08\%±1.64
&\textbf{90.55\%±1.12}
\\ % \hline
 MST-Prim
& 86.39\%±1.33
& 86.26\%±2.08
&\textbf{92.56\%±0.99}
\\  %\hline
 Topological Sort
& 87.27\%±2.67
& 81.65\%±2.53
&\textbf{87.30\%±4.62}
\\  %\hline
 Dijkstra
& 96.05\%±0.60
& 94.29\%±1.04
&\textbf{97.44\%±0.50}
\\  %\hline
 Binary Search
& 77.58\%±2.35
& 69.30\%±5.65
&\textbf{79.17\%±2.79}
\\  %\hline
 Bubble Sort
& 67.68\%±5.50
& 52.94\%±9.96
&\textbf{70.30\%±6.77}
\\ % \hline
 Graham Scan
& 93.62\%±0.91
& 87.74\%±3.87
&\textbf{94.58\%±0.87}
\\ % \hline
 Minimum
& 97.78\%±0.55
& 92.50\%±2.53
&\textbf{98.32\%±0.14}
\\  \hline
\end{tabular}
\end{table*}

%Moreover, we provide a heatmap in~\cref{xxx}.

%we show the highest-related task of each task in~\cref{xxx}.
% We select tasks with attention weights higher than the average $1/30$ for each task and present them in~\cref{xxx}.  

%From the table, we see \xcy{xxx heapsorts, min-heap greedy, task scheduling ,greedy sorting}.

%In CLRS, the 30 tasks are partitioned into 8 categories: Divide and Conquer, Dynamic Programming, Geometry, Graphs, Greedy, Search, Sorting, and Strings.

Surprisingly, the table indicates that the majority of tasks exhibit a preference for attention toward tasks outside their own categories, contrary to our initial expectations. Only four bolded pairs show high attention to tasks within the same category, with most of these being graph algorithms.
%For instance, the task that shows the greatest improvement by multi-task augmenting is the heapsort task, while the task it pays the most attention to is task scheduling, a greedy-type task. 
An intuitive explanation for this phenomenon is that tasks within the same category might not contribute additional information compared to the dataset used for training the task itself. Instead, tasks from other categories seem to play a crucial role in improving training accuracy. %For the heapsort task, the reasoning process involves maintaining a max or min heap, and this data structure shares some characteristics with greedy algorithms, leading to the task scheduling representation receiving significant attention.

%From the table, we observe that the majority of tasks exhibit higher attention within their respective categories, such as \xcy{Maybe not true xxx}. However, there are exceptions, such as the heapsort task, which shows the highest attention to task scheduling, which is a greedy-category task, with a calculated weight of \xcy{xxx}. An intuitive explanation is that the heapsort task involves maintaining a max or min heap, and this data structure shares some characteristics with greedy algorithms. Hence, the task scheduling representation receives significant attention from heapsort.

% This subsection delves into interpreting multi-task training. In our multi-task augmenting experiments, the acquired attention weights in the open-book processor reveal the significance of each task in relation to others. 

% For each task, we select tasks with attention weights higher than the average $1/30$ and present them in~\cref{xxx}. From the table, we see \cref{xxx}

%We selected the top five tasks based on their attention weights for each task and show them in~\cref{xxx}. 

%The above explanations offer some intuitive insights into multi-task training. Now, 

We proceed to a more in-depth examination of the relationships among tasks learned by the framework. We select a partner for each task according to~\cref{tab:attention} -- namely, the task it pays the most attention to. We conduct training and testing in a multi-task manner for each task paired with its chosen partner, and refer to this type of training as paired-task training. In this experiment, we only focus on tasks that either demonstrate accuracy improvements or slight declines in multi-task training compared to single-task training, and train them in the paired manner. The results are given in~\cref{tab:attention_verify}. The table validates our hypothesis. On these tasks, paired-task training achieves improvements compared to single-task training, with most tasks even surpassing the performance of multi-task training.

%and the results are 

%In the experiments, we select the tasks that yield accuracy improvements or slight declines in multi-task training compared to single-task training, and train them in the paired manner. The results are given in~\cref{tab:attention_verify}.

%that exhibit the greatest improvement under multi-task training. The training algorithm used was Triplet-GMPNN, and the experimental results are presented in~\cref{tab:attention_verify}. 

%For all these tasks, paired-task training achieves substantial improvements compared to single-task training. This indicates that the attention weights learned by our framework indeed reflect the inherent relationships among tasks.

%Specifically, we pick a pa

\subsection{Experimental Summary}\label{sec:analysis}

%\xcy{One table, all the results}

%We summarize the experiential results in~\cref{xxx}.

%\cref{tab:results} provides a summary of our test accuracies on CLRS. 

The experiments address the three questions posed at the beginning of the section. %We see the following trends from the experiments: 
%Due to space limitations, we use the column “Prior Best” to represent
%the best results among Memnet~\cite{DBLP:conf/icml/VelickovicBBPBD22}, PGN~\cite{DBLP:conf/icml/VelickovicBBPBD22}, MPNN~\cite{DBLP:conf/icml/VelickovicBBPBD22}. 

\begin{itemize}
    \item The open-book framework can significantly enhance the reasoning capabilities of various existing architectures, yielding state-of-the-art results across the majority of tasks in CLRS. 
    %This demonstrates the effectiveness of the open-book approach in enhancing the reasoning capabilities of various existing architectures.

    \item By feeding data from various tasks into the dataset encoder, the framework can successfully replicate the effects of multi-task training, and in most datasets, even outperform it.

    \item The attention-based implementation provides a valuable tool for the interpretability of multi-task training.  By examining the learned attention weights, we can gain insights into the influences and intrinsic relationships among tasks during multi-task training.

    % Checking the learned attention weights can help us figure out the influences and intrinsic relationships among tasks during multi-task training.
    
    % By feeding data from different tasks into the dataset encoder and allowing the open-book processor to learn the attention weights associated with these tasks, we can discern the influences and intrinsic relationships among tasks during multi-task training.

    % \item The open-book approach does not universally improve the network's reasoning abilities across all tasks. Our results indicate that, particularly in graph-type tasks, allowing the network to gather information from the training set can have a counterproductive effect. Tasks, where the open-book approach demonstrates positive effects, are mostly sorting or string-related tasks.

\end{itemize}

\section{Conclusion}\label{sec:con}

This paper considers open-book neural algorithmic reasoning, introducing a novel open-book framework accompanied by an attention-based implementation. Through empirical evaluations, we demonstrate that this implementation not only enhances the reasoning capabilities of the existing architecture but also functions as an effective tool for interpretable learning.

Several interesting direction for future research exist, such as exploring more effective implementations within the open-book framework. Note that although our current implementation demonstrates performance improvements for the majority of tasks in CLRS, there are instances where the open-book approach may yield counterproductive results. Refining the current architecture to ensure performance enhancements across all tasks remains a significant challenge.

%There are several interesting avenues for future research, e.g., exploring more effective implementations within the open-book framework. While our current implementation demonstrates performance improvements for the majority of tasks in CLRS, there are instances where employing the open-book approach may yield counterproductive results. Refining the current architecture to ensure performance enhancements across all tasks poses a substantial challenge.

%%%%%%%%%%%%%%%%%%%%%%%%%%%%%%%%%%%%%%%%%%%%%%%%%%%%%%%%%%%%
\newpage

\section*{ Acknowledgements}

This work is supported by the National Key Research Project of China under Grant No. 2023YFA1009402, the Scientific and Technological Innovation 2030 Major Projects under Grant 2018AAA0100902, NSFC Programs (62161146001, 62302166, 62372176), Shanghai Key Lab of Trustworthy Computing, Henan Key Laboratory of Oracle Bone Inscription Information Processing (AnYang Normal University), and the Key Laboratory of Interdisciplinary Research of Computation and Economics (SUFE), Ministry of Education.

\bibliographystyle{plain}
\bibliography{ref}

\newpage
\appendix

%\section{ Additional Experimental Results}\label{sec:add_exp}
%\iffalse
\section{Additional Experimental Results for Single-Task Augmenting}\label{sec:add_single}

\begin{figure*}[h]
  \centering
  \includegraphics[width=0.9\linewidth]{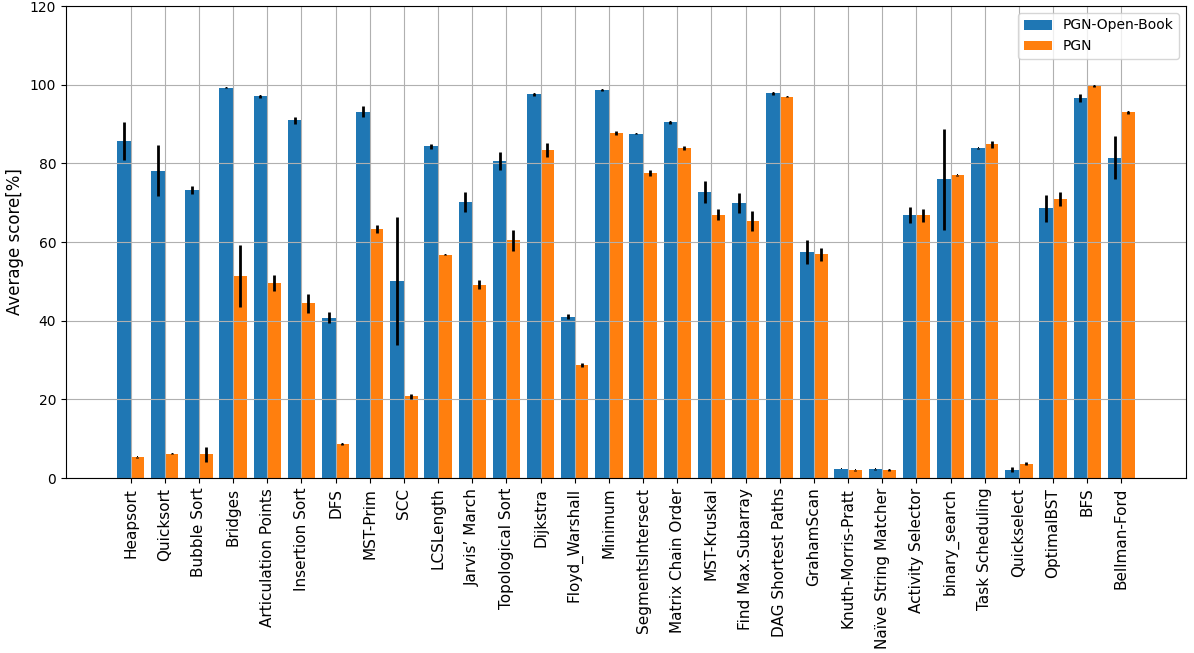}
  \caption{Comparison of the PGN architecture's performance before and after augmentation with the open-book framework. The 30 tasks are arranged in descending order of improvement magnitude.}
  \label{fig:PGN}
\end{figure*}

\begin{figure*}[h]
  \centering
  \includegraphics[width=0.9\linewidth]{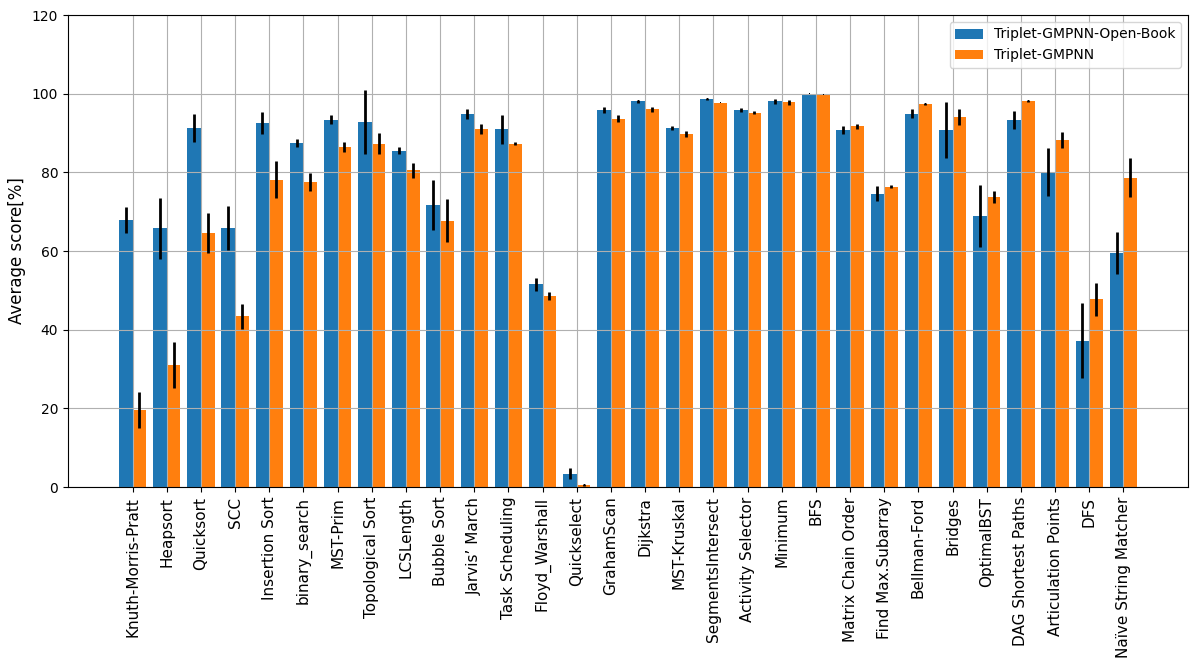}
  \caption{Comparison of the Triplet-GMPNN architecture's performance before and after augmentation with the open-book framework. The 30 tasks are arranged in descending order of improvement magnitude.}
  \label{fig:Triplet}
\end{figure*}

The comparison charts of PGN and Triplet-GMPNN are shown in~\cref{fig:PGN} and~\cref{fig:Triplet} respectively. The table that summarizes the results for all 30 tasks is present in~\cref{tab:results}.

\begin{table*}[h]
    \centering
    \caption{The summary of our obtained test accuracies on CLRS. The best-performing results in each row are highlighted in bold. The column ``Prior Best'' in the table represents the best results among four approaches in the literature: Memnet~\protect\cite{DBLP:conf/icml/VelickovicBBPBD22}, PGN~\protect\cite{DBLP:conf/icml/VelickovicBBPBD22}, MPNN~\protect\cite{DBLP:conf/icml/VelickovicBBPBD22}, and NPQ~\protect\cite{npq}, and the column ``Ours'' to denote the best results achieved by applying the open-book framework to the three existing architectures.}
    \label{tab:results}
    \begin{tabular}{cccc}
    \hline
        Task &
  %\begin{tabular}[c]{@{}c@{}}Prior Best\end{tabular} &
  \begin{tabular}[c]{@{}c@{}}Prior Best\end{tabular} &
  \begin{tabular}[c]{@{}c@{}}Triplet-GMPNN\end{tabular} &
  \multicolumn{1}{c}{\begin{tabular}[c]{@{}c@{}}Ours\end{tabular}}  \\ \hline
        Activity Selector   & 83.36\%±4.27 & 95.18\%±0.45 & \textbf{95.86\%±0.61}   \\ 
        Articulation Points   & 50.91\%±2.18 &88.32\%±2.01  & \textbf{98.30\%±0.35}   \\ 
        Bellman-Ford   & 92.99\%±0.34 & \textbf{97.39\%±0.19}  & 95.18\%±1.68     \\ 
        BFS   & 99.89\%±0.05 & 99.73\%±0.04  &\textbf{99.99\%±0.05}    \\ 
        Binary Search   & 76.95\%±0.13 & 77.58\%±2.35  & \textbf{87.44\%±1.07}    \\ 
        Bridges  & 72.69\%±4.78 & 93.99\%±2.07  & \textbf{99.26\%±0.04}   \\ 
        Bubble Sort   & 73.58\%±0.78 & 67.68\%±5.50  & \textbf{73.16\%±1.06}     \\
        DAG Shortest Paths   & 96.94\%±0.16 & \textbf{98.19\%±0.30}  & 97.79\%±0.34    \\ 
        DFS   & 13.36\%±1.61 & \textbf{47.79\%±4.19}  & 40.79\%±1.44    \\ 
        Dijkstra   & 91.50\%±0.50 & 96.05\%±0.60  & \textbf{98.29\%±0.42}     \\ 
        Find Max.Subarray   & 65.23\%±2.56 & \textbf{76.36\%±0.43}  & 74.52\%±1.88     \\ 
        Floyd-Warshall   & 28.76\%±0.51 &48.52\%±1.04  & \textbf{51.52\%±1.74}     \\ 
        Graham Scan  & 91.04\%±0.31 & 93.62\%±0.91  & \textbf{96.19\%±0.06}    \\ 
        Heapsort   & 68.00\%±1.57 & 31.04\%±5.82 & \textbf{85.71\%±4.82}    \\ 
        Insertion Sort   & 71.42\%±0.86 & 78.14\%±4.64  & \textbf{92.61\%±2.82}     \\ 
        Jarvis’March   & 92.88\%±2.87 & 91.01\%±1.30  & \textbf{94.74\%±1.27}    \\
        Knuth-Morris-Pratt   & 3.91\%±0.15 & 19.51\%±4.57  & \textbf{71.24\%±3.7}    \\ 
        LCS Length   & 72.05\%±5.72 & 80.51\%±1.84  & \textbf{85.54\%±0.87}   \\ 
        Matrix Chain Order   & 83.91\%±0.49 & \textbf{91.68\%±0.59}  & 90.85\%±1.11   \\ 
        Minimum   & 87.71\%±0.52 & 97.78\%±0.55  & \textbf{98.65\%±0.27}    \\ 
        MST-Kruskal   & 70.97\%±1.50 & 89.80\%±0.77  & \textbf{91.26\%±0.56}     \\ 
        MST-Prim   & 69.08\%±7.56 & 86.39\%±1.33  & \textbf{93.41\%±0.52}     \\
        Naive String Matcher   & 4.24\%±0.98 & \textbf{78.67\%±4.99}  & 73.57\%±1.62     \\
        Optimal BST   & 72.03\%±1.21 & \textbf{73.77\%±1.48}  & 70.04\%±2.38    \\
        Quickselect  & \textbf{3.66\%±0.42} & 0.47\%±0.25  & 3.37\%±1.37     \\ 
        Quicksort   & 73.10\%±0.67 & 64.64\%±5.12  & \textbf{83.13\%±3.52}    \\ 
        Segments Intersect   & 93.53\%±0.88 & 97.64\%±0.09  & \textbf{98.71\%±0.16}     \\ 
        SCC   & 32.19\%±9.23 & 43.43\%±3.15  & \textbf{65.83\%±5.51}  \\ 
        Task Scheduling   & 84.89\%±0.91 & 87.25\%±0.35  & \textbf{90.93\%±3.63}     \\
        Topological Sort   & 60.45\%±2.69 & 87.27\%±2.67  & \textbf{92.80\%±8.09}     \\ \hline
        Overall Average  & 66.04\% & 75.98\% & \textbf{82.91\%}  \\  \hline
        % \textgreater 90\% & 7/30 & 11/30 & a & b & c  \\ 
        % \textgreater 80\% & 11/30 & 17/30 & a & b & c  \\ 
        % \textgreater 60\% & 23/30 & 24/30 & a & b & c  \\ \hline
    \end{tabular}
\end{table*}

\clearpage
\section{Additional Experimental Results for Multi-Task Interpretation}\label{sec:add_multi}

\begin{figure}[tb]
  \centering
  \includegraphics[width=0.8\linewidth]{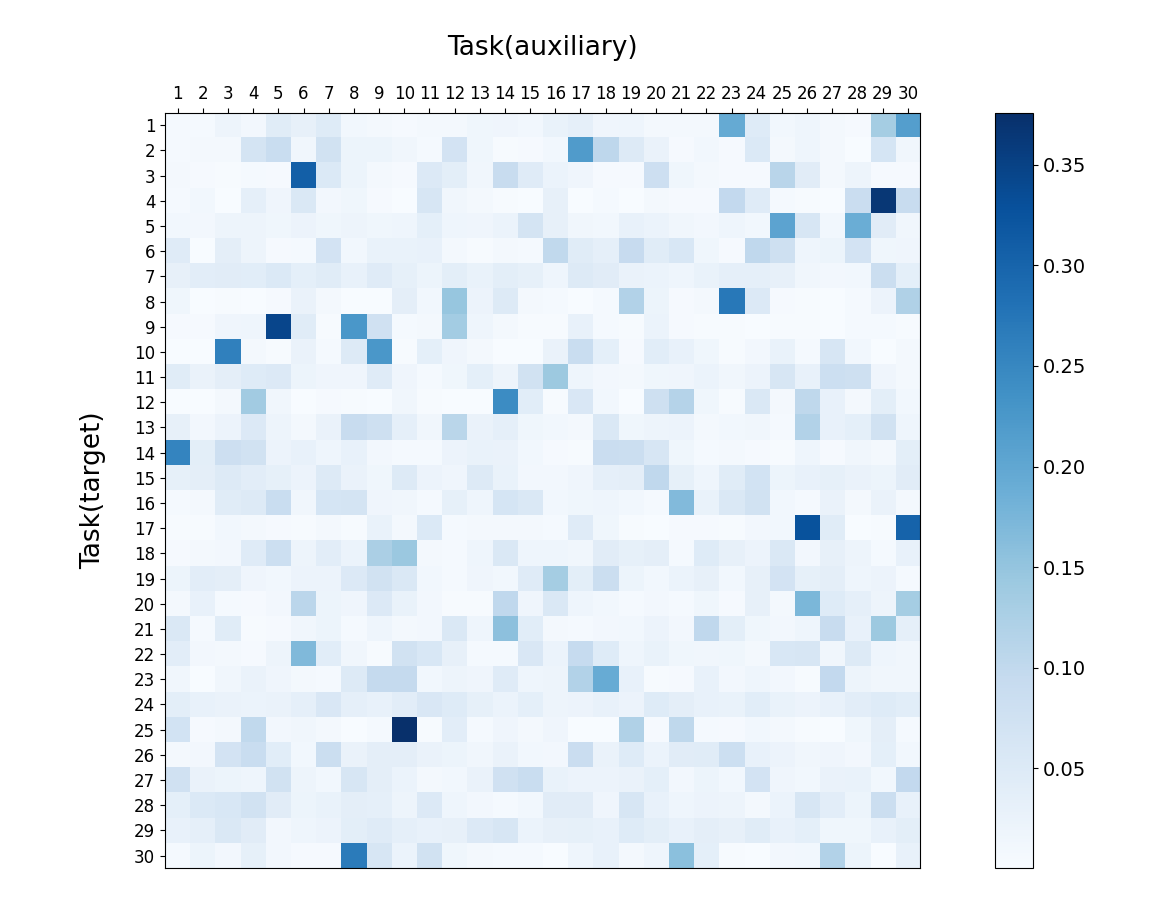}
  \caption{A heatmap where each row represents the attention vector of a (target) task in CLRS. }
  \label{fig:heat}
\end{figure}

We redraw the table that shows the learned task relationships and assign an index to each task in the new table (\cref{tab:attention_app}). Using the indices, we present a heatmap regarding the attention weights in~\cref{fig:heat}.

\begin{table*}[htbp]
\caption{For each (target) task, we show the task with the highest attention weight among other tasks in column ``Auxiliary Task''. The last column indicates whether the two tasks belong to the same category, displaying $\surd$ if they do.}
\label{tab:attention_app}
\centering
\begin{tabular}{cccc}
\hline
 &
  \begin{tabular}[c]{@{}c@{}}Target Task\end{tabular} &
  \begin{tabular}[c]{@{}c@{}}Auxiliary Task\end{tabular} &
  \begin{tabular}[c]{@{}c@{}}Category Membership\end{tabular} \\ \hline
1  & Activity Selector & Topological Sort & \\ \hline
2  & Articulation Points & Knuth-Morris-Pratt& \\ \hline
3  & Bellman-Ford & Bridges& \\ \hline
4  & BFS & Task Scheduling& \\ \hline
5  & Binary Search & Quickselect & $\surd$\\ \hline
6  & Bridges & Optimal BST & \\ \hline
7  & Bubble Sort & Task Scheduling& \\ \hline
8  & DAG Shortest Paths & Naïve String Matcher & \\ \hline
9  & DFS & Binary Search & \\ \hline
10 & Dijkstra & Bellman-Ford & $\surd$\\ \hline
11 & Find Max. Subarray & Jarvis’ March& \\ \hline
12 & Floyd-Warshall & Heapsort& \\ \hline
13 & Graham Scan & Quicksort & \\ \hline
14 & Heapsort & Activity Selector & \\ \hline
15 & Insertion Sort & Minimum & \\ \hline
16  & Jarvis’ March & MST-Kruskal & \\ \hline
17  & Knuth-Morris-Pratt & Quicksort & \\ \hline
18  & LCS Length & Dijkstra & \\ \hline
19  & Matrix Chain Order & Jarvis’ March& \\ \hline
20  & Minimum & Quicksort & \\ \hline
21  & MST-Kruskal & Heapsort & \\ \hline
22  & MST-Prim & Bridges & $\surd$\\ \hline
23  & Naïve String Matcher & LCS Length & \\ \hline
24  & Optimal BST & Find Max. Subarray & \\ \hline
25 & Quickselect & Dijkstra& \\ \hline
26 & Quicksort & BFS& \\ \hline
27 & Segments Intersect & Topological Sort & \\ \hline
28 & SCC& Task Scheduling& \\ \hline
29 & Task Scheduling& Heapsort & \\ \hline
30 & Topological Sort & DAG Shortest Paths & $\surd$\\ \hline
\end{tabular}
\end{table*}

\clearpage
\section{Ablation Study for The Number of Auxiliary Data Points }\label{sec:ablation}

\begin{figure*}[htbp]
  \centering
  \includegraphics[width=0.8\linewidth]{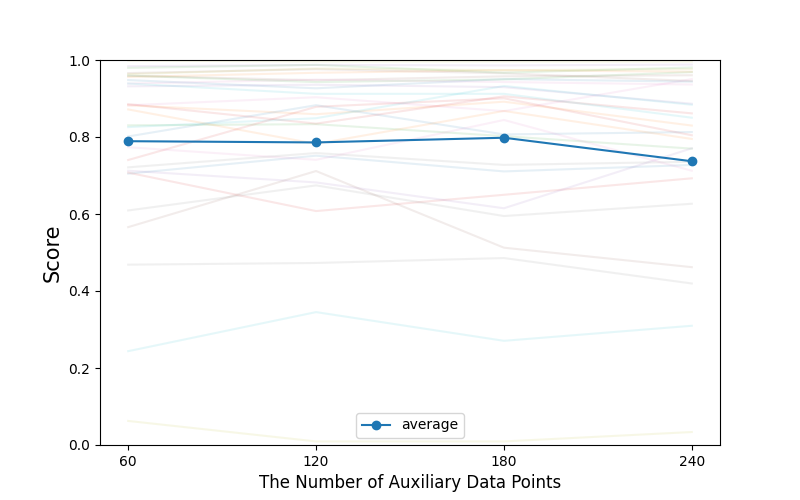}
  \caption{The summary of our results when the number of auxiliary data points varies from 60 to 240. We blur the performance curves for each task, with the solid curve representing the average value.}
  \label{fig:ablation_auxiliary}
\end{figure*}

This section presents the performance of our framework when the number of auxiliary data points varies.~\cref{fig:ablation_auxiliary} demonstrates the robustness. 

% \begin{table*}[h]
%     \centering
%     \caption{The summary of our results on each task category when the number of auxiliary data points varies from 60 to 240.}
%     \label{tab:8sort_abla}
%     \begin{tabular}{ccccc}
%     \hline
%         Task Category &
%   %\begin{tabular}[c]{@{}c@{}}Prior Best\end{tabular} &
%   60 &
%   120 &
%   180 & 240 \\ \hline
%         Graphs   &80.54\%±2.39 &82.57\%±2.23  &80.99\%±2.05 & \textbf{85.37\%±1.73}   \\ 
%         Geometry   &96.07\%±0.36 &96.12\%±0.63  &96.21\%±0.82 & \textbf{96.55\%±0.50}   \\ 
%         Dynamic Programming   &80.67\%±2.42 &76.67\%±2.24  &78.13\%±0.77 & \textbf{82.14\%±1.45}    \\ 
%         Divide and Conquer   &70.48\%±2.86 &75.15\%±2.27  &71.06\%±2.90 & 74.52\%±1.88    \\ 
%         Greedy & 88.68\%±4.07& 88.76\%±1.61 &94.15\%±0.80 & \textbf{93.40\%±2.12}   \\ 
%         Search   &92.55\%±1.88 &88.48\%±2.05  & 91.73\%±1.79& \textbf{63.15\%±0.90}     \\
%         Sorting   &70.35\%±10.96 & 70.69\%±11.74 &73.50\%±6.83 & \textbf{83.65\%±3.06}    \\ 
%           \hline
%         % \textgreater 90\% & 7/30 & 11/30 & a & b & c  \\ 
%         % \textgreater 80\% & 11/30 & 17/30 & a & b & c  \\ 
%         % \textgreater 60\% & 23/30 & 24/30 & a & b & c  \\ \hline
%     \end{tabular}
% \end{table*}

\clearpage
\section{Scalability of Our Framework}\label{sec:scalability}

%This section investigates 

This section further investigates the scalability of our framework. 
For each dataset in the CLRS benchmark, the training set graphs contain approximately 12 nodes on average and the test set graphs contain 64 nodes.
We design two additional experiments (in the context of single-task augmenting). Note that due to memory constraints caused by the increased graph size, we omit the string category tasks and the quickselect algorithm tasks in these two experiments.

\begin{figure*}[htbp]
  \centering
  \includegraphics[width=0.8\linewidth]{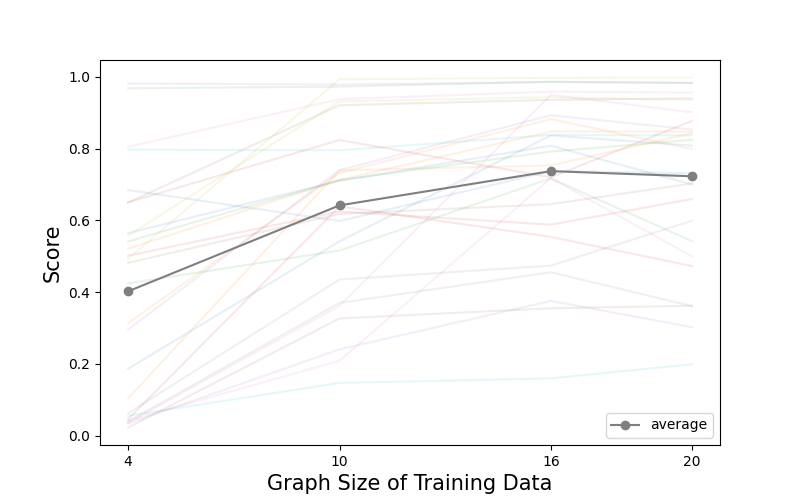}
  \caption{The results of training data scaling.}
  \label{fig:training_scale}
\end{figure*}

\begin{figure*}[htbp]
  \centering
  \includegraphics[width=0.8\linewidth]{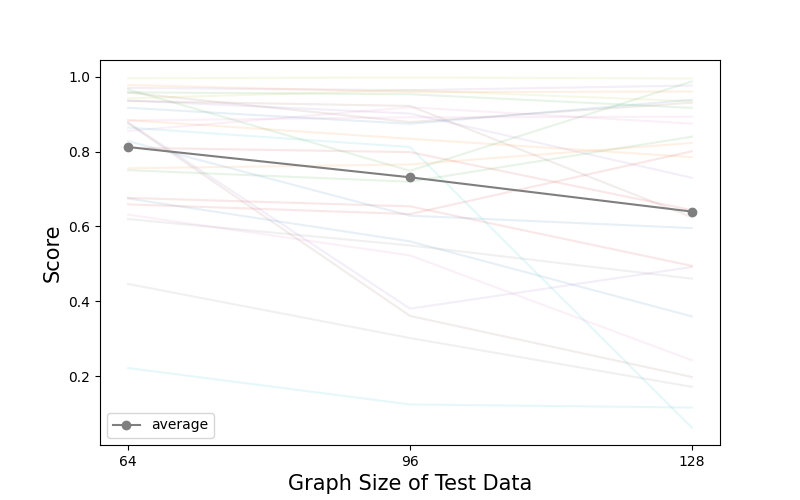}
  \caption{The results of test data scaling.}
  \label{fig:test_scale}
\end{figure*}

\begin{itemize}
    \item \textbf{Training Data Scaling:} In this experiment, we fix the original testing graph size and vary the graph size in the training data set from $4$ to $20$. The results are present in~\cref{fig:training_scale}.

    \item \textbf{Test Data Scaling:} In this experiment, we fix the original training graph size and vary the graph size in the test data set from $64$ to $128$. The results are present in~\cref{fig:test_scale}.
\end{itemize}

From the figures, we can see that as the graph size of the training data and test data varies, our approach maintains robust out-of-distribution performance.

\end{document}